\documentclass[10pt,letterpaper]{article}
\usepackage[T1]{fontenc}
\usepackage{lmodern}
\usepackage[top=0.85in,left=2.75in,footskip=0.75in]{geometry}
\usepackage{amsmath,amssymb,amsfonts}
\usepackage{changepage}
\usepackage{textcomp,marvosym}
\usepackage{cite}
\usepackage{nameref,hyperref}
\usepackage[nopatch=eqnum]{microtype}
\DisableLigatures[f]{encoding = *, family = * }
\usepackage[table]{xcolor}
\usepackage{array}
\usepackage{booktabs}
\newcolumntype{+}{!{\vrule width 2pt}}
\newlength\savedwidth

\usepackage{setspace}
\raggedright
\makeatletter
\renewcommand\paragraph{\@startsection{paragraph}{4}{\z@}%
  {3.25ex \@plus1ex \@minus.2ex}%
  {6pt \@plus2pt \@minus2pt}%
  {\normalfont\normalsize\bfseries}}
\makeatother
\usepackage[aboveskip=1pt,labelfont=bf,labelsep=period,justification=raggedright,singlelinecheck=off]{caption}

\makeatletter
\renewcommand{\@biblabel}[1]{\quad#1.}
\makeatother
\usepackage{lastpage,fancyhdr,graphicx}
\usepackage{epstopdf}
\fancyheadoffset[L]{2.25in}
\fancyfootoffset[L]{2.25in}
\usepackage{acronym}
\usepackage{listings}
\graphicspath{{./}}

\acrodef{AI}{artificial intelligence}
\acrodef{LLM}{Large Language Model}
\acrodef{NLP}{natural language processing}
\acrodef{IRR}{inter-rater reliability}
\acrodef{PA}{Percent Agreement}
\acrodef{PABAK}{Prevalence-Adjusted Bias-Adjusted Kappa}
\acrodef{CI}{confidence interval}
\acrodef{QA}{question answering}

\begin{document}

\begin{flushleft}
{\Large
Evaluating Large Language Model Raters for German Open-Response Clinical Questions: A Physician-Annotated Benchmark Study of Agreement, Evaluator Bias, and Abstention
}
\newline
\\
William Philipp, MSc\textsuperscript{1,*},
Finn Fassbender, BSc\textsuperscript{2,4},
Daniel Fister\textsuperscript{3,4},
Thorsten Langer, Prof.\ Dr.\ med.\textsuperscript{5},
Martje G. Pauly, Dr.\ med.\textsuperscript{6},
Rebecca Herzog, Dr.\ med.\textsuperscript{6},
Markus A.\ Hobert, Dr.\ med.\textsuperscript{6},
Theresa Paulus, Dr.\ med.\textsuperscript{6},
Alexander Baumann, Dr.\ med.\textsuperscript{7},
Chi Wang Ip, Prof.\ Dr.\ med.\textsuperscript{8},
Lukas L. Goede, Dr.\ med.\textsuperscript{9},
Johanna Engelhard, Dr.\ med.\textsuperscript{9},
Sebastian Loens, Dr.\ med.\textsuperscript{10},
Ronald B\"{o}ck, Prof. Dr.-Ing.\textsuperscript{11},
Sebastian Fudickar, Prof.\ Dr. rer. nat.\textsuperscript{1,12}
\\
\bigskip
1 Section for Clinical Research-IT, Institute of Medical Biometry and Statistics, University of Luebeck, Luebeck, Germany
\\
2 Medical Faculty, University of T\"{u}bingen, T\"{u}bingen, Germany
\\
3 Medical Faculty, Martin Luther University Halle-Wittenberg, Halle, Germany
\\
4 German Medical Students' Association (bvmd), Berlin, Germany
\\
5 Department of Pediatric and Adolescent Medicine, University of Luebeck / University Hospital Schleswig-Holstein, Luebeck, Germany
\\
6 Department of Neurology, University of Luebeck, Luebeck, Germany
\\
7 Department of Neurology, University of Kiel, Kiel, Germany
\\
8 Department of Neurology, University Hospital W\"{u}rzburg, W\"{u}rzburg, Germany
\\
9 Department of Neurology, Charit\'{e} -- Universit\"{a}tsmedizin Berlin, Berlin, Germany
\\
10 Institute of Systems Motor Science, University of Luebeck, Luebeck, Germany
\\
11 Faculty of Digital Administration, Meissen University of Applied Administrative Sciences (UAAS), Meissen, Germany
\\
12 Fraunhofer Research Institution for Individualized Medical Technology and Engineering (IMTE), Luebeck, Germany
\\
\bigskip
* w.philipp@uni-luebeck.de

\end{flushleft}

\clearpage
\newgeometry{top=0.85in,left=1in,right=1in,footskip=0.75in}

\section*{Abstract}

\textbf{Background:} \acp{LLM} increasingly support clinical language tasks, but their evaluation on non-English open-response questions is constrained by a lack of expert-annotated benchmarks. \acs{LLM}-as-a-judge evaluators could make such assessments scalable, but their suitability for clinical evaluation requires validation.

\textbf{Objective:} To introduce MedQADE, a standardized German open-response clinical benchmark with physician reference annotations, and to evaluate \acs{LLM}-as-a-judge systems for alignment with physician judgments, self- and intra-family evaluation bias, and abstention behavior.

\textbf{Methods:} MedQADE comprises 3,800 question-answer sets, each containing answers from five student \acp{LLM} and reference annotations from a panel of ten physicians. The 200-question core was rated by all ten raters; the 3,600-question extension was assigned two primary raters per question, with the tenth physician resolving disagreements. We applied nine \ac{LLM} evaluators to all sets and assessed physician reliability, student-model accuracy, alignment with physician consensus, evaluator bias, and abstention.

\textbf{Results:} Physicians showed moderate-to-substantial agreement on answer correctness, with an unweighted mean pairwise Cohen's~$\kappa$ of 0.612, but limited agreement on question difficulty (Krippendorff's~$\alpha = 0.208$ with squared distances between numeric difficulty scores). Student-model accuracy ranged from 17.8\% to 66.0\% and generally decreased with physician-rated difficulty. The strongest evaluator, Gemini 3 Flash, approached the leave-one-out physician reference ($\kappa = 0.694$ versus 0.709). Paired analyses identified these biases: four of five models rated their own responses more favorably than out-of-family evaluators, and five of six same-family comparisons were positive. Physician abstention increased with perceived difficulty, whereas seven of nine \ac{LLM} evaluators abstained in no more than 0.51\% of evaluations; two of the strongest evaluators assigned definitive labels to every response.

\textbf{Conclusions:} MedQADE supports physician-anchored evaluation of German open-response clinical answers. Strong \ac{LLM} evaluators can approach physician agreement in this setting, but evaluator bias and low observed abstention warrant physician validation and further assessment of selective deferral before fully automated benchmark evaluation. These results do not establish clinical safety.

\noindent\textbf{Keywords:} artificial intelligence; large language models; medical education; clinical natural language processing; question answering; benchmarking; automated evaluation; German language.

\section*{Introduction}\label{sec:intro}

\subsection*{Clinical NLP and the Need for Benchmarking}

Clinical \ac{NLP} supports medical workflows, such as referral-letter triage, patient-journey processing, and \ac{AI}-assisted screening of clinical referrals \cite{Fudickar2024, Klug2024, Maarseveen2025}. The corresponding extraction and structuring of relevant information from routine clinical texts can improve  tasks such as prioritization and care coordination in practice \cite{Fudickar2024, Maarseveen2025}. With the emergence of \acfp{LLM}, the scope of clinical language processing has expanded beyond traditional rule-based approaches toward a wider range of clinical applications including summarization, generation, question answering, and patient-facing communication tasks \cite{Busch2025, Oliveira2025, Mandal2025}. As \acp{LLM} become more widely integrated into clinical practice, their reliable evaluation and appropriate regulatory oversight become increasingly important.

Clinical \ac{AI} applications may fall under the high-risk provisions of the EU \ac{AI} Act and may also be subject to the Medical Device Regulation (MDR). The MDR mandates clinical evaluation based on sufficient clinical evidence for devices within its scope \cite{MDR2017}. Therefore, suitable evaluation of clinical \acp{LLM} is increasingly important. However, to characterize the diverse outputs and behaviors of modern \acp{LLM}, conventional task-specific metrics and evaluation methods are often insufficient, since they must account for the variability in generated texts and differences in model performance across languages. Reliable and reproducible benchmarks are therefore needed to systematically compare model capabilities and to support the validation of clinical \ac{AI} systems.

\subsection*{Limitations of Current Medical Benchmarks}

Current medical \ac{LLM} evaluation relies heavily on selected-response, particularly multiple-choice formats. English resources such as PubMedQA \cite{Jin2019}, MedMCQA \cite{Pal2022}, and MedQA \cite{Jin2021} have played an important role in benchmarking biomedical and medical language models. Similar approaches have also been adopted for other languages, including Swedish \cite{Moell2025}, Polish \cite{Rosol2023}, and Chinese \cite{Luo2025}.

Although Germany has the highest healthcare expenditure in the EU (per a 2023 survey) \cite{WozniakHolecka2026}, German-language resources remain limited. GerMedIQ provides a German corpus of simulated anamnesis question-response pairs \cite{Hofenbitzer2025}. More recently, DeFineMed \cite{Doll2026} employed machine-translated versions of MMLU \cite{Hendrycks2021} and MedQA \cite{Jin2021} to evaluate specialized German medical language models, explicitly identifying the reliance on translated data as a limitation resulting from the lack of publicly available, expert-curated German medical \ac{QA} datasets. Despite this need for native German evaluation resources, large-scale native German clinical knowledge benchmarks with expert reference annotations are, to our knowledge, still unavailable.

The reliance on selected-response formats (such as multiple-choice tests) introduces a more fundamental limitation. Multiple-choice tests measure whether a model can identify the correct answer among provided alternatives, whereas many clinical tasks require independently retrieving, formulating, and synthesizing relevant information. In human assessment, free recall is generally more demanding than recognition-based testing \cite{Rowland2014}, and open-response formats can provide information about competence that is not captured by selected-response formats \cite{Procop2026, Sheaffer2013}. Recent evidence suggests that the same distinction is relevant for \acp{LLM}. Singh et al. found that medical \ac{LLM} performance decreased by 39.43\% when questions were changed from multiple-choice to an open-response format \cite{Singh2025}. More broadly, recent work has shown that high multiple-choice performance can mask sensitivity to answer format and other superficial characteristics of the benchmark \cite{Cocchieri2026}. Thus, performance on selected-response questions does not necessarily translate into the ability to independently produce clinically appropriate answers or support open-ended reasoning in clinical practice.

\subsection*{Open-Response Evaluation}

Open-response evaluation addresses this limitation by requiring models to generate an answer without being presented with candidate solutions. Such evaluation is increasingly being explored in medical \ac{LLM} benchmarking.
MORQA combines multilingual medical open-response QA with multiple expert-authored reference answers and expert ratings, and shows that \ac{LLM}-based evaluation metrics can better capture agreement with human judgments than traditional lexical metrics \cite{MORQA2025}. MediQAl extends an examination-derived medical benchmark with 4,969 open-response questions alongside multiple-choice tasks \cite{Bazoge2026}.

However, the flexibility of open-response questions introduces an additional evaluation challenge. Whereas the correctness of a multiple-choice response can usually be determined from a discrete label, an open response can take many forms while still being clinically correct. Responses may differ in wording, level of detail, ordering, or terminology, and may contain both correct and incorrect information. Simple string matching or other rule-based metrics are therefore insufficient for many clinical open-response tasks.

Expert physician judgments remain important for establishing and validating reference assessments of open-ended medical responses, but manual annotation can be costly and difficult to scale. This creates a practical bottleneck for systematic comparisons of modern \acp{LLM}: removing answer options broadens the assessment task while increasing the need for judgments that accommodate clinically acceptable variations in responses.

\subsection*{LLM-as-a-Judge}

The \ac{LLM}-as-a-Judge paradigm offers an automated alternative to manual evaluation, enabling open-response outputs to be evaluated at scale while potentially reducing the physician annotation workload. It leverages \acp{LLM} as judges to assess the quality of another model's response \cite{Chiang2023}.  The judge receives the original question, the candidate response, and either a reference answer or an evaluation rubric, and is prompted to produce a structured assessment of correctness, completeness, or other task-specific criteria.

Recent work has shown that automated judges can achieve robust agreement with expert physicians  \cite{Croxford2025} and, in some cases, produce a more stable consensus than individual human raters \cite{Chen2025}. Recent clinical benchmarking work has further demonstrated the feasibility of combining expert-developed evaluation criteria with automated judging and validating the resulting scores against human assessments \cite{Zhang2025}. While the \ac{LLM}-as-a-Judge paradigm may reduce manual annotation demands, the judge as part of the evaluation system must itself be validated. Validation against physician assessments is therefore important before using automated judges to compare medical \acp{LLM} at scale. For evaluation, the following two dimensions stand out:

\begin{itemize}
\item \ac{LLM} evaluators may exhibit self-enhancement bias, especially when scoring their own outputs \cite{Bowman2024, Ackerman2024}. Such biases could distort model comparisons even when overall agreement with human judgments appears high.
\item Although medical \acp{LLM} have been shown to frequently fail to abstain under clinical uncertainty \cite{Cocchieri2026Abstain}, it remains unclear whether an automated judge can recognize uncertainty and abstain when it lacks a sufficient basis for a reliable judgment.
\end{itemize}

\subsection*{Study Objectives and Contributions}

The primary objective of this study is to investigate whether \acp{LLM} can serve as reliable automated annotators for German open-response clinical questions. Establishing such reliability requires both a physician reference standard and an evaluation framework that distinguishes agreement from systematic evaluator behavior. We therefore address five research questions:

\begin{description}
    \setlength{\itemsep}{0pt}
    \setlength{\parskip}{0pt}
    \setlength{\parsep}{0pt}
    \item[RQ1:] How consistently do physicians judge the correctness of student-\ac{LLM} answers?
    \item[RQ2:] How well do student \acp{LLM} perform on German open-response clinical questions?
    \item[RQ3:] How does individual and ensemble \ac{LLM}-evaluator alignment compare with physician reliability?
    \item[RQ4:] Do \ac{LLM} evaluators show bias when rating their own or same-family responses compared with out-of-family evaluators rating the same responses?
    \item[RQ5:] How do physician and \ac{LLM}-evaluator abstention patterns compare?
\end{description}

By answering these questions, this study makes three main contributions:

\begin{itemize}
\item We introduce the MedQADE benchmark, a German open-response clinical benchmark comprising 3,800 physician-annotated question-answer sets and 19,000 answer slots sampled from a filtered pool of 26,598 German medical examination-style questions. The associated research materials comprise the question text, physician annotations, generated answers, reference labels, automated-evaluator outputs, and analysis code.
\item We quantify physician reliability and the performance of five student \acp{LLM}, characterizing the consistency of the physician reference and the performance range represented in the benchmark.
\item We evaluate nine \acp{LLM} as automated evaluators against the physician reference standard along three dimensions: alignment with physician judgments, evaluator bias, and abstention behavior. These analyses identify limitations that can affect the reliability of fully automated benchmark evaluation.
\end{itemize}

\section*{Methods}\label{sec:methods}

\begin{figure}[!p]
\centering
\includegraphics[width=0.875\textwidth]{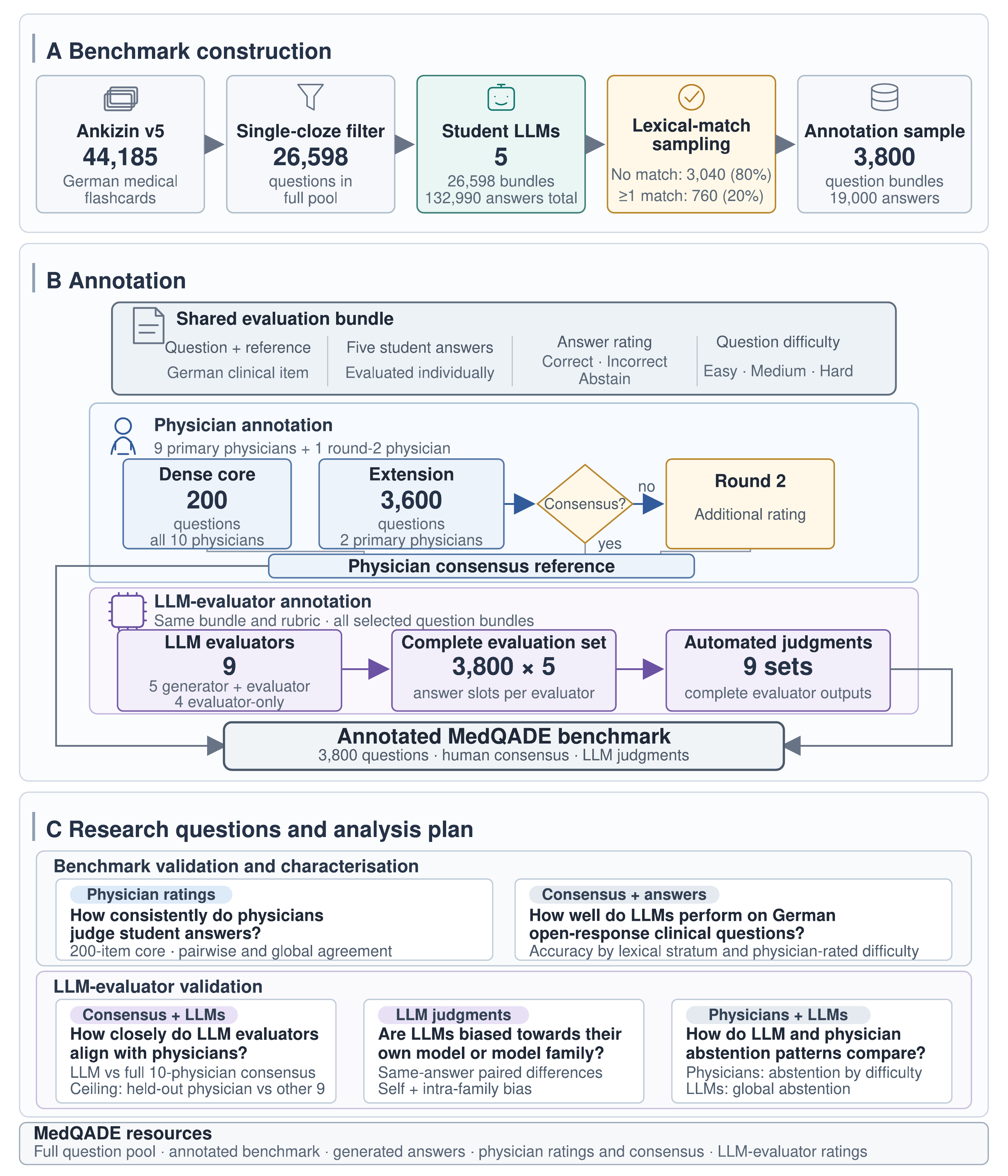}
\caption{MedQADE benchmark construction, annotation, and analysis framework. (A) Five student large language models (LLMs) answered single-cloze questions derived from a collection of German medical flashcards. Stratified sampling based on lexical agreement with the reference answer yielded 3,800 five-answer question-answer sets. (B) Physicians and nine LLM evaluators assessed the same question-answer sets using a shared rubric covering answer correctness, abstention, and question difficulty. The 200-question core was rated by all ten raters. Extension questions were assigned two primary raters; the tenth physician provided additional ratings when consensus was absent. Physician ratings and consensus labels form the benchmark annotations; LLM judgments constitute evaluation data. (C) Analyses examine physician reliability, student-LLM performance, LLM–physician alignment, evaluator bias, and physician-versus-LLM abstention patterns. The dataset includes the complete 26,598-question pool and generated answers. GPT-5.6 Sol assisted with creation of this overview figure; the authors reviewed its final content.}
\label{fig:overview}
\end{figure}

\subsection*{Study Design}\label{sec:study_design}

The study comprised two phases (Figure~\ref{fig:overview}). First, we constructed MedQADE by preprocessing a source corpus, generating answers with five student \acp{LLM}, selecting question-answer sets for annotation, and establishing physician reference labels. Second, we used the resulting benchmark to quantify physician reliability and student-model performance and to evaluate nine \ac{LLM} annotators with respect to alignment, evaluator bias, and abstention behavior. Benchmark construction and the automated-evaluator protocol are described below, followed by research-question-specific quantitative analyses. Because this was a benchmark-construction and evaluator-validation study rather than a clinical trial, diagnostic-accuracy study, or patient-level prediction-model study, no established design-specific reporting checklist directly applied.

\subsection*{MedQADE Benchmark Construction}

\subsubsection*{Source Corpus and Preprocessing}

The source question pool for MedQADE was derived from version 5 of the Ankizin project \cite{Ankizin2024}, a digital flashcard collection for Anki, an open-source spaced-repetition platform widely used by German medical students preparing for the German Second State Examination in Medicine. Although developed as a study tool, the collection provided a suitable foundation for our benchmark because of its scale, grounding in medical education, quality-control procedures, and open-answer format. It contains German-language clinical questions and is actively maintained by medical students and domain experts. Each flashcard undergoes peer review for medical accuracy before inclusion \cite{AnkizinLeitfaden2024}. All questions follow a cloze-style, open-answer format, in contrast to comparable medical question-answering datasets, which typically use multiple-choice questions.

The raw Ankizin v5 dataset contains 44,185 flashcards. We converted the native Anki export into a standardized format and retained only items containing exactly one cloze deletion. This filtering produced a source pool of 26,598 single-cloze items. An English translation of a representative retained item is shown below:

\begin{quote}
\small
Question: What is the trigger for Guillain-Barr\'{e} syndrome (GBS) in $\frac{2}{3}$ of cases? \hrulefill\\
Target Answer: Previous infections
\end{quote}

\subsubsection*{Synthetic Student Answer Generation}

To investigate RQ 1 and 2, we used five \acp{LLM} as proxies for medical students to answer all 26,598 items: Gemini 2.5 Flash \cite{GeminiTeam2025a}, GPT-5 Nano \cite{OpenAI2025}, Gemma 3 27B \cite{GemmaTeam2025}, Gemma 3 4B \cite{GemmaTeam2025}, and Qwen3-4B \cite{QwenTeam2025}. Exact implementation identifiers and execution details are documented in \hyperref[S1_Configuration]{Multimedia Appendix 1}; the student prompt is reproduced in \hyperref[S2_Appendix]{Multimedia Appendix 2}. The model selection allowed comparisons between proprietary and open-weight models, while the two Gemma variants provided a within-family comparison across parameter scales. The proprietary models were accessed through their official APIs, whereas the open-weight models were run locally using Ollama, an \ac{LLM} serving framework.

Each model received a German-language system prompt designed to approximate a testing environment (see \hyperref[S2_Appendix]{Multimedia Appendix 2}). The prompt instructed the model to assume the role of an examinee completing a time-critical, high-stakes medical examination, building on previous work on role-play and emotion-oriented prompting \cite{Kong2024, Li2023}.

Student generation used model-specific temperatures and the lowest available reasoning or thinking settings documented in \hyperref[S1_Configuration]{Multimedia Appendix 1}; GPT-5 Nano used the provider's default temperature because user-defined values were not supported.

\subsubsection*{Sampling Stratified by Lexical Match}\label{sec:methods_sampling}

3,800 items were selected for annotation from the filtered 26,598 items using lexical-match-based stratification: We defined a binary sampling variable based on exact lexical agreement between the generated and reference answers. Items were classified as \textit{exact lexical match} if at least one of the five student models reproduced the reference answer verbatim under strict string equality; all other items were classified as \textit{no exact lexical match}. We expected items in the former stratum to yield higher answer accuracy, whereas the latter would contain a greater proportion of generated answers that were incorrect or expressed non-verbatim.

Among the 3,800 items, we applied a 20:80 allocation of 760 \textit{exact lexical match} and 3,040 \textit{no exact lexical match} items. This allocation enriched the benchmark for items whose generated answers were expected to be incorrect or non-verbatim while retaining a comparison stratum with direct lexical agreement.

\subsubsection*{Physician Annotation Task and Guidelines}

Human annotation was done via a custom interface in the Label Studio platform, with items showing the clinical question, the five generated student answers, and the original Ankizin reference answer. The reference answer was provided as guidance.

Raters completed two annotation tasks:

\begin{enumerate}
\item Categorical correctness: Each student answer was assigned one of three labels:
\begin{itemize}
\item \textit{Correct}: The answer was medically accurate and sufficiently complete, without false statements. Variations in phrasing or level of detail were acceptable if the core medical meaning was preserved.
\item \textit{Incorrect}: The answer contained factual errors, omitted information necessary for a complete response, or was too vague to establish correctness. This category also included technically inaccurate or potentially unsafe medical statements.
\item \textit{Abstain}: The rater lacked the specific medical expertise required to judge the answer confidently.
\end{itemize}

\item Subjective difficulty: Raters classified each question as \textit{Easy}, \textit{Medium}, or \textit{Hard} according to its perceived clinical difficulty. This rating was used only for downstream statistical stratification and did not affect the correctness labels assigned to the generated answers.
\end{enumerate}

The complete annotation guidelines, including clinical examples and resolutions for edge cases, are available in \hyperref[S3_Appendix]{Multimedia Appendix 3}.

\subsubsection*{Physician Annotation and Reference-Standard Construction}\label{sec:methods_physician_reference}

The 3,800 items were assigned to nine primary physician raters using a split annotation design. Within this sample, we randomly selected a shared core of 200 items while preserving the 20:80 lexical-match allocation. The core was independently rated by all ten raters. The remaining 3,600 items were distributed among the primary physicians using a deterministic round-robin offset, with two primary raters assigned per item. Each primary physician was assigned 1,000 items: the 200-item shared core and 800 items from the split allocation. Analyses used available parseable ratings; missing or unparseable ratings were excluded. Both item assignment and presentation order were randomized.

To create physician consensus labels, consensus for each student answer was defined as a simple majority among the available non-abstaining ratings. For the split-annotation items, the tenth physician independently evaluated all five answers and question difficulty when the initial ratings lacked a simple majority for at least one answer. Consensus was then recomputed from all available non-abstaining ratings, with the tenth physician's ratings included as additional votes. All physicians annotated independently, without discussion or calibration between panel members.

The resulting MedQADE benchmark comprises 3,800 physician-annotated question-answer sets (a 200-item dense core and a 3,600-item set), each containing five generated answers (19,000 answer slots).
Each benchmark record contains the question, original reference answer, student-model outputs and identities, physician correctness and difficulty ratings, and consensus correctness labels.
The complete filtered pool of 26,598 questions and all generated answers are part of the question text and evaluation data described in Data Availability.
The automated-evaluator judgments described below are experimental outputs included in the evaluation data rather than components of the physician reference standard.

\subsection*{Automated \ac{LLM} Evaluator Protocol}

After constructing the physician-annotated benchmark, we applied nine language models as automated evaluators to all 3,800 question-answer sets. Each model independently rated the five generated answers and question difficulty using the same label set and medical criteria as the physician panel. The evaluator panel included the five student models: Gemini 2.5 Flash \cite{GeminiTeam2025a}, Gemma 3 4B \cite{GemmaTeam2025}, Gemma 3 27B \cite{GemmaTeam2025}, GPT-5 Nano \cite{OpenAI2025}, and Qwen3-4B \cite{QwenTeam2025}. It additionally included Gemini 3 Flash \cite{GeminiTeam2025b}, Gemini 2.5 Flash-Lite \cite{GeminiTeam2025a}, GPT-5.4 Nano \cite{OpenAI2026}, and GPT-5.4 Mini \cite{OpenAI2026}. Exact evaluator identifiers, provider aliases, and approximate recorded run-completion times are documented in \hyperref[S1_Configuration]{Multimedia Appendix 1}. Including models not represented in the student pool enabled assessment of evaluator bias by comparing their ratings with those from out-of-family evaluators.

Evaluator temperature was set to 0 where supported; GPT-5 Nano used the provider's default temperature because user-defined values were not supported. The lowest available reasoning or thinking settings were selected, with model-specific settings documented in \hyperref[S1_Configuration]{Multimedia Appendix 1}.

The prompts retained the medical criteria used for human annotation while incorporating the following components:

\begin{itemize}
\item Role-play prompting: Each model was assigned the role of an expert medical evaluator, aligning the task framing with that of the physician panel \cite{Kong2024}.
\item Rubric-based prompting: The prompts reproduced the human criteria for \textit{Correct}, \textit{Incorrect}, and \textit{Abstain}. Explicit categorical rubrics have been reported to improve the consistency of \ac{LLM}-based evaluation \cite{Kim2023}.
\item Pointwise evaluation: All five answers were included in the same prompt to mirror the human annotation interface, but the model was instructed to assess each answer independently against the rubric rather than rank the answers against one another.
\item Structured output: Responses were constrained using a JSON Schema containing the predefined correctness and difficulty labels. This facilitated automated parsing and prevented outputs outside the permitted label set.
\end{itemize}

The full evaluator rubric and prompt are reproduced in \hyperref[S3_Appendix]{Multimedia Appendix 3}.

\subsection*{Ethical Considerations}
The unit of analysis was a clinical question-answer set and the study evaluated model-generated responses and expert annotations of nonidentifiable educational materials. All physician annotators were study investigators and coauthors who contributed professional assessments; individual ratings supported the reliability and abstention analyses, but results were reported as panel summaries rather than identifiable evaluations of individual physicians. The study involved no patients, clinical care or clinical decision-making, intervention, participant recruitment, qualitative interviews, biobank or registry data, or patient-identifiable information. As such, no ethics committee approval was sought or obtained.

\subsection*{Quantitative Analyses}

\subsubsection*{Physician Reliability (RQ1)}\label{sec:methods_irr}

On the 200-item core subset, we assessed the reliability of the categorical correctness labels using raw \ac{PA}, Cohen's~$\kappa$, and \ac{PABAK}. Let $n$ denote the number of available paired ratings for which neither physician abstained, $n_{\mathrm{agree}}$ the number of these pairs with matching labels, $p_o$ the observed agreement, and $p_e$ the agreement expected from the physicians' marginal \textit{Correct} and \textit{Incorrect} proportions. For each physician pair and answer slot,
\[
\mathrm{PA}=p_o=\frac{n_{\mathrm{agree}}}{n},
\qquad
\kappa=\frac{p_o-p_e}{1-p_e},
\qquad
\mathrm{PABAK}=2p_o-1.
\]
The three metrics were calculated for all 45 unique physician pairs and averaged without weighting within each answer slot. Cohen's~$\kappa$ adjusts for agreement expected by chance, whereas \ac{PABAK} was included because it is less sensitive to prevalence and marginal bias effects \cite{Byrt1993}. Cohen's~$\kappa$ values were interpreted using the Landis and Koch scale \cite{Landis1977}.

To assess all ten physicians jointly while allowing missing ratings, we additionally calculated Krippendorff's~$\alpha=1-D_o/D_e$, where $D_o$ and $D_e$ denote observed and expected disagreement. Correctness used nominal disagreement and was analyzed separately for each answer slot.

Nonparametric 95\% confidence intervals were estimated by bootstrapping the 200 core questions. In each of 1,000 iterations, 200 questions were sampled with replacement while retaining all associated ratings from the ten physicians. Krippendorff's~$\alpha$, mean pairwise Cohen's~$\kappa$, and mean pairwise \ac{PA} were then recalculated separately for each answer slot. The 2.5th and 97.5th percentiles of the resulting bootstrap distributions defined the confidence intervals. Missing ratings remained missing during resampling and were handled according to the valid-rating rules of each metric. The \ac{PABAK} confidence limits were obtained by applying the same linear transformation to the \ac{PA} confidence limits.

For each correctness metric, the overall value was calculated as the unweighted mean of the five slot-specific point estimates; an overall confidence interval was not calculated. The bootstrap intervals therefore quantify uncertainty arising from the sampled questions conditional on the participating physician panel.

Reliability of the subjective difficulty ratings was assessed on the same core subset using Krippendorff's~$\alpha$ with squared distances between numeric difficulty scores. The categories were encoded as $\textit{Easy}=0$, $\textit{Medium}=1$, and $\textit{Hard}=2$, and the disagreement distance between scores $x$ and $y$ was $(x-y)^2$. This numeric-score distance differs from the standard frequency-dependent ordinal distance. Consequently, an \textit{Easy}--\textit{Hard} disagreement had four times the weight of an adjacent-category disagreement.

\subsubsection*{Student-Model Performance (RQ2)}\label{sec:methods_student}

Student-model accuracy was calculated against the physician consensus labels. For consensus-dependent analyses, answer slots without a resolved physician consensus were treated as missing. Items were assigned to four equally spaced tiers of mean physician-rated question difficulty: $[0,0.5]$, $(0.5,1.0]$, $(1.0,1.5]$, and $(1.5,2.0]$.

Accuracy per student model was defined as the proportion of its responses labeled \textit{Correct}. Two-sided 95\% confidence intervals for the overall and stratified accuracy estimates were calculated using the Wilson score method with $z=1.96$. Accuracy was stratified by both the pre-annotation lexical-match variable and the mean physician-rated difficulty tiers. The former analysis assessed whether the sampling strata differentiated model performance, whereas the latter examined how accuracy varied with physician-perceived difficulty. Comparing these stratifications enabled the assessment of the observed association between lexical-match status and subsequent physician difficulty ratings.
For the reported between-stratum accuracy contrasts, 95\% percentile confidence intervals were calculated from 5,000 independent-subgroup bootstrap iterations resampling question rows within each stratum.

\subsubsection*{\ac{LLM} Evaluator Alignment and Ensembles (RQ3)}\label{sec:methods_llm_alignment}

Alignment between the \ac{LLM} annotators and the ten-physician consensus was assessed on the 200-item core subset. For each annotator, \ac{PA}, Cohen's~$\kappa$, and \ac{PABAK} were calculated separately for each of the five answer slots and then averaged without weighting. Annotator abstentions and unresolved physician consensus labels were excluded pairwise.

Because agreement between a single rater and an aggregated consensus is not directly comparable to pairwise agreement between individual physicians, we calculated a leave-one-out physician reference. Each physician $r$ was held out in turn, and a leave-one-out consensus $C^{(-r)}$ was constructed separately for each answer slot from the available non-abstaining ratings of the other nine physicians. A strict majority was required; ties and cases without substantive ratings remained unresolved. The held-out physician was compared with $C^{(-r)}$ using \ac{PA}, Cohen's~$\kappa$, and \ac{PABAK}. Comparisons excluded abstentions. Each metric was averaged first across the five answer slots for each physician and then across the ten physicians.

For both the physician reference and each \ac{LLM} comparison, 95\% confidence intervals were estimated using 1,000 question-level bootstrap iterations that preserved all five answer slots within each sampled question.

We additionally conducted an exploratory, in-sample combinatorial search for majority-vote \ac{LLM} ensembles. Gemma 3 4B and Qwen3-4B were excluded because their individual alignment suggested that they were unlikely to improve majority-vote performance. The resulting pool comprised Gemini 3 Flash, Gemini 2.5 Flash, Gemini 2.5 Flash-Lite, GPT-5.4 Mini, GPT-5.4 Nano, GPT-5 Nano, and Gemma 3 27B. All possible subsets of sizes 1, 3, 5, and 7 were evaluated on the core subset. Abstentions were removed before majority voting. Ensemble predictions were treated as missing when all members abstained, whereas post-abstention ties were assigned \textit{Incorrect}. Separate optimal ensembles were selected by maximizing $\kappa$, \ac{PA}, and \ac{PABAK}, respectively. Because both candidate screening and ensemble selection were performed on the same core subset, the resulting estimates represent exploratory, in-sample alignment.

\subsubsection*{Evaluator Bias (RQ4)}\label{sec:methods_llm_bias}

We assessed self- and intra-family evaluation bias by testing whether, on responses generated by their own model or by another model from the same architectural model family, \ac{LLM} annotators assigned more favorable ratings than did out-of-family annotators for the complete 3,800-item dataset. The architectural model-family groups were Gemini, Gemma, GPT, and Qwen; Gemini and Gemma were treated as separate families despite sharing Google as their provider. We used a response-level paired comparison: for each response, the target annotator's score was compared with the mean score assigned by out-of-family annotators to the same response. Because both scores refer to identical response content and the same clinical question, their difference cannot be attributed to variation in student performance or question difficulty. Using the mean of several out-of-family annotators also reduced the influence of idiosyncratic ratings from any single comparison model.

We first quantified self-evaluation bias. Let $A$ denote the target annotator, $R_{\text{self}}$ the set of eligible responses generated by the same model as $A$, and $S(a,r_i) \in \{0,1\}$ the score assigned by annotator $a$ to response $r_i$, with 1 representing \textit{Correct} and 0 representing \textit{Incorrect}. For each response $r_i$, let $A_{\text{peer}}$ denote the annotators outside $A$'s architectural model family that provided a non-abstaining rating. Their mean score was defined as

\[
\bar{S}_{\text{peer}}(r_i) = \frac{1}{|A_{\text{peer}}|} \sum_{a \in A_{\text{peer}}} S(a, r_i)
\]

This bias was defined as the mean paired difference between the target annotator's score and the out-of-family mean for the same responses:

\[
\Delta_{\text{self}}(A) = \frac{1}{|R_{\text{self}}|} \sum_{r_i \in R_{\text{self}}} \left[ S(A, r_i) - \bar{S}_{\text{peer}}(r_i) \right]
\]

A positive value indicates that the target model rated its own responses more favorably than did the out-of-family annotators, whereas a negative value indicates less favorable self-evaluation.

The same paired comparison was used to assess intra-family evaluation bias on same-family responses. Let $m$ denote a sibling student model from the same architectural model family as $A$, but not the identical model, and let $R_m$ denote the eligible responses generated by that sibling. The corresponding paired difference was defined as

\[
\Delta_{\text{family}}(A,m) = \frac{1}{|R_m|} \sum_{r_i \in R_m} \left[ S(A, r_i) - \bar{S}_{\text{peer}}(r_i) \right]
\]

Each target-annotator--sibling-model pairing was analyzed separately and responses from different sibling models were not pooled into a single family estimate. Abstentions and missing evaluator records were excluded from the target scores and the corresponding peer means.

Uncertainty was quantified using 1,000 bootstrap iterations. Questions were sampled with replacement, with each sampled question contributing the paired difference for the relevant student-answer slot. The 2.5th and 97.5th percentiles of the bootstrapped mean differences defined the 95\% confidence intervals for $\Delta_{\text{self}}(A)$ and each $\Delta_{\text{family}}(A,m)$.

\subsubsection*{Physician and \ac{LLM} Abstention (RQ5)}

We analyzed physician abstention separately within each difficulty tier. For each physician and tier, the abstention rate was defined as the number of answer-slot ratings labeled \textit{Abstain} divided by the total number of non-missing answer-slot ratings provided by that physician. The resulting per-physician rates were used to examine the association between perceived item difficulty and abstention behavior.

For each \ac{LLM} evaluator, we calculated both the global abstention rate and abstention rates within the same four physician-derived difficulty tiers used for the physician analysis. Within each tier, the rate was defined as the number of answer slots labeled \textit{Abstain} divided by all answer-slot ratings provided by that evaluator. Because every \ac{LLM} evaluated all five responses for every question, the four tier-specific denominators were 9,010, 6,330, 2,860, and 800 ratings, respectively. Difficulty ratings assigned by the \acp{LLM} were not used for this stratification.

\section*{Results}\label{sec:results}

\subsection*{Physician Reliability (RQ1)}

Physician agreement on answer correctness was moderate to substantial across the five student models (Table~\ref{tab:human_irr}). Mean pairwise Cohen's~$\kappa$ ranged from 0.555 to 0.656; the unweighted means for $\kappa$ and Krippendorff's~$\alpha$ were 0.612 and 0.598, respectively. Although mean \ac{PA} was 80.1--89.1\%, its divergence from $\kappa$ was greatest for Gemma 3 4B and Qwen3-4B, for which only 13.6\% and 16.6\% of resolved responses were classified as \textit{Correct}, respectively.

\begin{table}[!ht]
\caption{Physician inter-rater reliability for answer correctness. PA denotes percent agreement; PABAK denotes prevalence-adjusted bias-adjusted kappa; $\kappa$ denotes Cohen's kappa; and $\alpha$ denotes Krippendorff's alpha.}
\label{tab:human_irr}
\begin{adjustwidth}{-0.65in}{-0.65in}
\centering
\scriptsize
\setlength{\tabcolsep}{3pt}
\begin{tabular}{@{}lcccc@{}}
\toprule
Student model & Cohen's~$\kappa$ & \acs{PA} (\%) & \acs{PABAK} & Krippendorff's~$\alpha$ \\
\midrule
Gemini 2.5 Flash & 0.555 [0.494, 0.613] & 80.1 [77.2, 82.9] & 0.601 [0.544, 0.659] & 0.548 [0.486, 0.604] \\
Gemma 3 4B       & 0.618 [0.514, 0.711] & 89.1 [86.3, 91.7] & 0.783 [0.726, 0.834] & 0.599 [0.502, 0.690] \\
Gemma 3 27B      & 0.656 [0.595, 0.713] & 82.9 [79.9, 85.7] & 0.659 [0.598, 0.714] & 0.642 [0.576, 0.700] \\
GPT-5 Nano       & 0.587 [0.526, 0.646] & 80.4 [77.5, 83.3] & 0.609 [0.551, 0.666] & 0.578 [0.517, 0.637] \\
Qwen3-4B         & 0.644 [0.556, 0.721] & 87.9 [85.2, 90.4] & 0.757 [0.703, 0.808] & 0.625 [0.538, 0.701] \\
\midrule
Unweighted mean  & 0.612 & 84.1 & 0.682 & 0.598 \\
\bottomrule
\end{tabular}
\end{adjustwidth}
\vspace{2pt}
\raggedright\scriptsize Values are point estimates [95\% confidence interval] for the 200-item core subset. Pairwise metrics were averaged across the 45 physician pairs for each answer slot; Krippendorff's~$\alpha$ used ratings from the ten-physician panel. Confidence intervals were obtained by question-level bootstrap. The unweighted means do not have confidence intervals.
\end{table}

In the extension, the additional physician rating resolved 2,537 of 2,770 initially unresolved answer slots (91.6\%). The remaining 233 unresolved slots and 120 newly tied slots left 353 of 18,000 extension slots without consensus; together with 21 core slots, this yielded 374 of 19,000 slots (2.0\%) excluded from consensus-dependent analyses. Agreement on question difficulty in the core subset was limited (Krippendorff's~$\alpha=0.208$ with squared distances between numeric difficulty scores) and was substantially lower than agreement on answer correctness.

\subsection*{Student-Model Performance (RQ2)}

Student-model accuracy ranged from 17.8\% to 66.0\% overall, and every model performed better on items with an exact lexical match to the reference answer (Table~\ref{tab:student_lexical}). The observed exact-match advantage ranged from 9.8 to 33.3 percentage points; for Gemini 2.5 Flash, it was 33.3 percentage points (95\% percentile bootstrap CI 30.7-36.0).

\begin{table}[!ht]
\caption{Student-model accuracy by lexical-match stratum.}
\label{tab:student_lexical}
\begin{adjustwidth}{-0.55in}{-0.55in}
\centering
\scriptsize
\setlength{\tabcolsep}{4pt}
\begin{tabular}{@{}lcccc@{}}
\toprule
Student model & Overall accuracy & Exact match & No exact match & Difference (pp) \\
\midrule
Gemini 2.5 Flash & 66.0 [64.4, 67.5] & 92.5 [90.4, 94.1] & 59.2 [57.4, 61.0] & 33.3 \\
GPT-5 Nano       & 56.6 [55.0, 58.2] & 76.2 [73.0, 79.1] & 51.6 [49.8, 53.4] & 24.6 \\
Gemma 3 27B      & 43.1 [41.5, 44.7] & 63.3 [59.8, 66.6] & 37.9 [36.2, 39.7] & 25.3 \\
Qwen3-4B         & 19.2 [18.0, 20.5] & 28.1 [25.0, 31.4] & 17.0 [15.7, 18.4] & 11.2 \\
Gemma 3 4B       & 17.8 [16.6, 19.1] & 25.6 [22.6, 28.9] & 15.9 [14.6, 17.2] & 9.8 \\
\bottomrule
\end{tabular}
\end{adjustwidth}
\vspace{2pt}
\raggedright\scriptsize Accuracy values are percentages [two-sided 95\% Wilson confidence interval] among responses with a resolved physician consensus. Exact-match status indicates whether at least one student model reproduced the reference answer exactly. Differences are exact-match minus no-exact-match accuracy in percentage points (pp). Resolved answer-slot denominators (overall/exact/no match), in table order: 3,720/757/2,963; 3,704/752/2,952; 3,720/757/2,963; 3,737/754/2,983; 3,745/753/2,992. Totals before exclusions were 3,800/760/3,040 per model.
\end{table}

\begin{figure}[!ht]
\centering
\includegraphics[width=\linewidth]{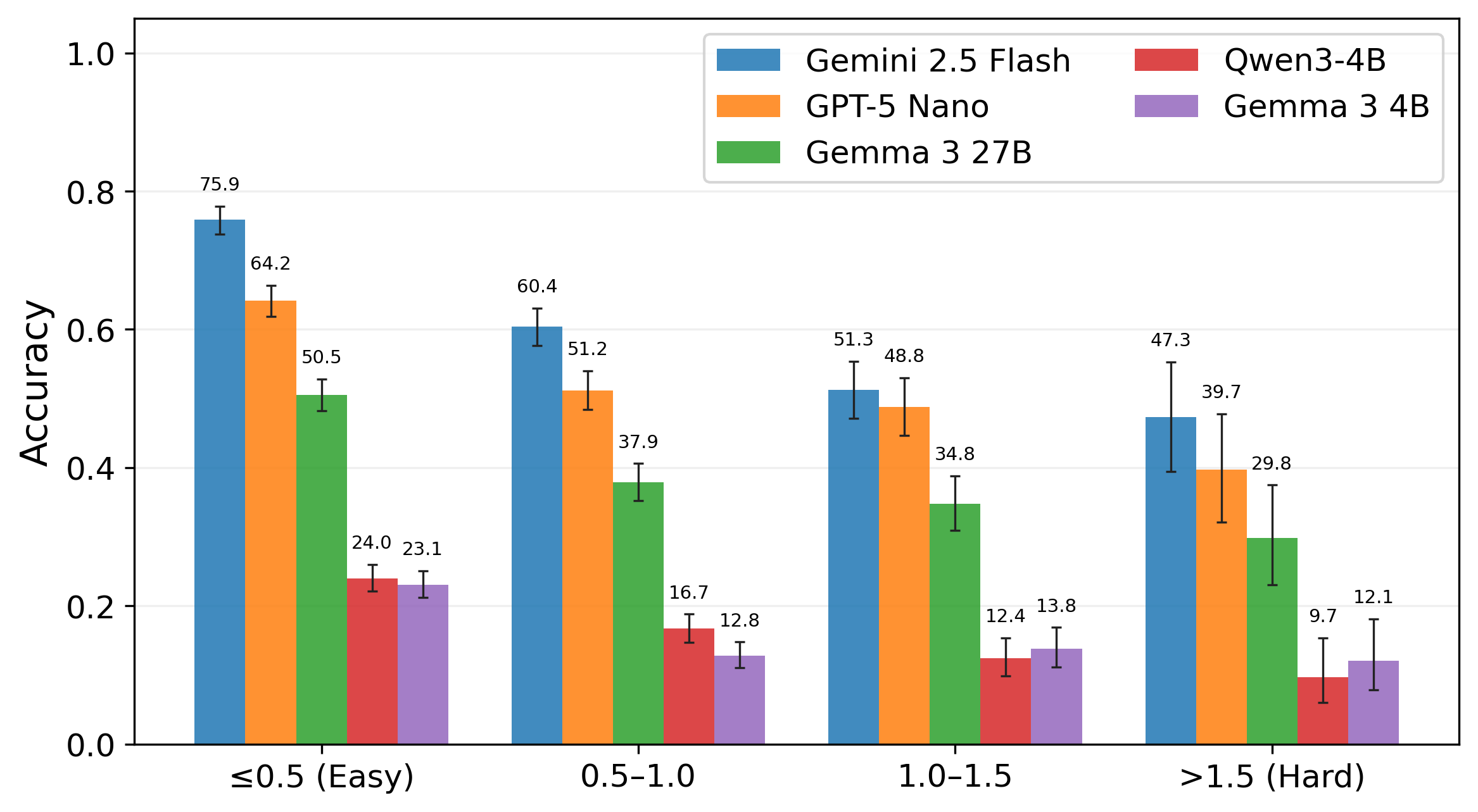}
\caption{Student-model accuracy by physician-rated difficulty. Accuracy across four tiers of mean physician-rated difficulty. The tiers contained 1,802, 1,266, 572, and 160 questions, respectively. Bars show the proportion of responses labeled \textit{Correct} among those with a resolved physician consensus; error bars indicate two-sided 95\% Wilson score confidence intervals. Per-model resolved answer-slot denominators ranged from 1,771 to 1,783, 1,231 to 1,243, 548 to 564, and 146 to 157 across the four tiers, respectively.}
\label{fig:student_difficulty}
\end{figure}

Lexical-match status partially tracked physician-rated difficulty: exact-match items constituted 27.1\% (489/1,802) of the lowest difficulty tier and 14.4\% (182/1,266), 12.8\% (73/572), and 10.0\% (16/160) of the successive tiers. Accuracy was generally lower at greater difficulty (Figure~\ref{fig:student_difficulty}); for Gemini 2.5 Flash, it was 75.9\% (1,353/1,783 resolved answer slots) in the lowest tier versus 47.3\% (70/148) in the highest, a difference of 28.6 percentage points (95\% percentile bootstrap CI 20.5-36.8).

\subsection*{\ac{LLM} Evaluator Alignment and Ensembles (RQ3)}

Gemini 3 Flash was closest to the leave-one-out physician reference on the 200-item core subset, with $\kappa=0.694$ versus 0.709 and overlapping confidence intervals for all three agreement metrics (Table~\ref{tab:llm_alignment}). GPT-5.4 Mini also overlapped the physician reference for all three metrics; Gemini 2.5 Flash-Lite overlapped only for $\kappa$, whereas Gemini 2.5 Flash overlapped only for \ac{PA} and \ac{PABAK}. The two 4B evaluators had the lowest $\kappa$ values: 0.444 for Qwen3-4B and 0.327 for Gemma 3 4B.

\begin{table}[!ht]
\caption{Large language model (LLM) evaluator alignment with physician consensus. PA denotes percent agreement; PABAK denotes prevalence-adjusted bias-adjusted kappa; and $\kappa$ denotes Cohen's kappa.}
\label{tab:llm_alignment}
\begin{adjustwidth}{-0.35in}{-0.35in}
\centering
\scriptsize
\setlength{\tabcolsep}{4pt}
\begin{tabular}{@{}lccc@{}}
\toprule
Evaluator & Cohen's~$\kappa$ & \acs{PA} (\%) & \acs{PABAK} \\
\midrule
Physician reference       & 0.709 [0.668, 0.746] & 88.3 [86.7, 89.8] & 0.766 [0.733, 0.796] \\
\midrule
Gemini 3 Flash            & \textit{0.694 [0.619, 0.754]} & \textit{87.8 [84.9, 90.3]} & \textit{0.756 [0.697, 0.807]} \\
GPT-5.4 Mini              & \textit{0.616 [0.544, 0.682]} & \textit{84.5 [81.5, 87.5]} & \textit{0.691 [0.629, 0.749]} \\
Gemini 2.5 Flash-Lite     & \textit{0.602 [0.520, 0.674]} & 83.0 [79.5, 86.1] & 0.659 [0.591, 0.723] \\
Gemini 2.5 Flash          & 0.595 [0.523, 0.667] & \textit{83.6 [80.3, 86.8]} & \textit{0.672 [0.606, 0.736]} \\
GPT-5.4 Nano              & 0.560 [0.478, 0.638] & 82.1 [78.7, 85.5] & 0.643 [0.574, 0.711] \\
Gemma 3 27B               & 0.558 [0.479, 0.625] & 81.0 [77.6, 84.2] & 0.620 [0.552, 0.684] \\
GPT-5 Nano                & 0.541 [0.460, 0.617] & 82.0 [78.8, 85.1] & 0.641 [0.575, 0.703] \\
Qwen3-4B                  & 0.444 [0.365, 0.520] & 78.8 [75.6, 82.2] & 0.576 [0.511, 0.643] \\
Gemma 3 4B                & 0.327 [0.252, 0.403] & 69.2 [65.5, 73.1] & 0.383 [0.310, 0.462] \\
\bottomrule
\end{tabular}
\end{adjustwidth}
\vspace{2pt}
\raggedright\scriptsize Values are point estimates [question-level bootstrap 95\% confidence interval]. The physician reference is the mean agreement between each held-out physician and the consensus of the other nine physicians; each \acs{LLM} was compared with the ten-physician consensus. Italicized LLM cells indicate metric-specific confidence-interval overlap with the corresponding physician reference. Interval overlap is descriptive and does not establish equivalence.
\end{table}

The selected five-model ensemble---Gemini 3 Flash, Gemini 2.5 Flash, GPT-5 Nano, GPT-5.4 Nano, and Gemma 3 27B---had unweighted macro-$\kappa=0.6947$ versus 0.6936 for Gemini 3 Flash alone (exploratory in-sample difference $+0.0011$, 95\% paired item-bootstrap percentile CI $-0.057$ to $+0.058$); candidate screening and ensemble selection used the same core subset. Gemini 3 Flash alone remained optimal for \ac{PA} (87.8\%) and \ac{PABAK} (0.756).

\subsection*{Evaluator Bias (RQ4)}

\begin{figure}[!ht]
\centering
\includegraphics[width=\linewidth]{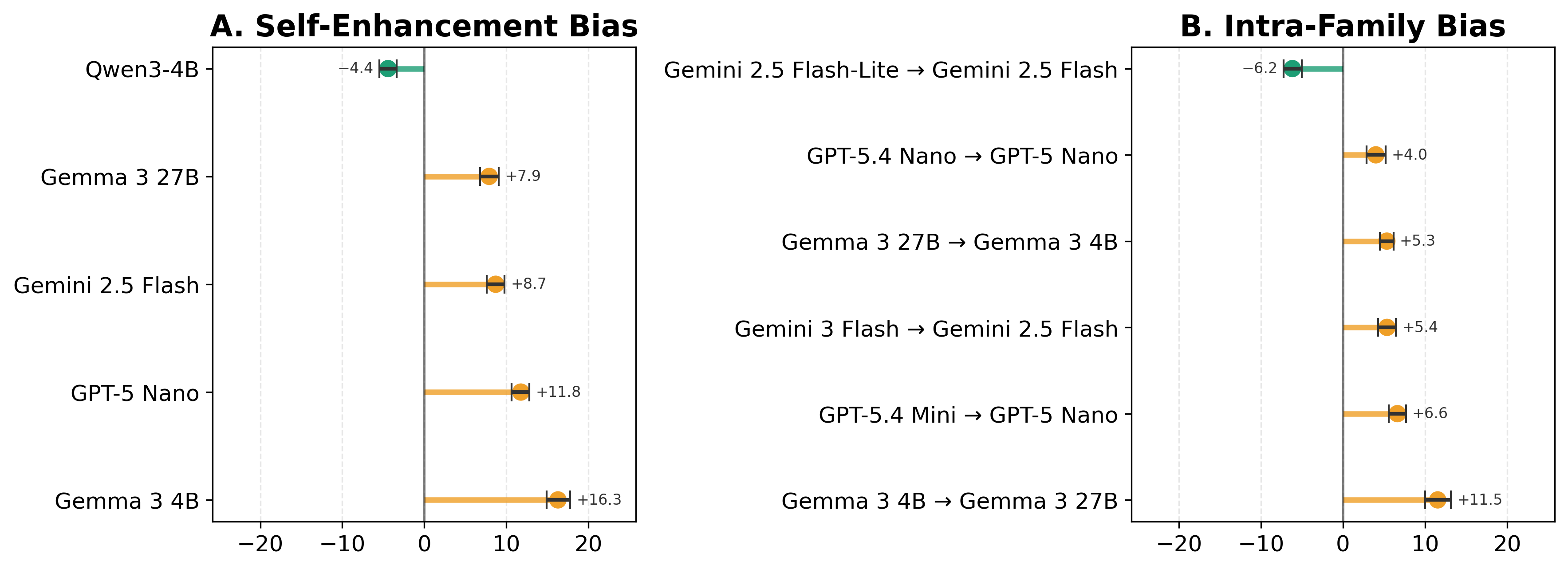}
\caption{Bias in large language model (LLM) evaluators. Mean paired differences between each target annotator's correctness ratings and the mean ratings assigned by out-of-family annotators to the same responses. Values are expressed as percentage points. Panel A shows responses generated by the same model as the annotator; Panel B shows responses generated by another model from the same architectural model family. Families were Gemini, Gemma, GPT, and Qwen, with Gemini and Gemma treated separately. Positive values indicate more favorable ratings by the target annotator, whereas negative values indicate less favorable ratings. Error bars indicate question-level bootstrap 95\% confidence intervals. The notation \textit{Model A} $\rightarrow$ \textit{Model B} denotes Model A evaluating responses generated by Model B.}
\label{fig:evaluation_bias}
\end{figure}

These biases were observed in both comparison sets (Figure~\ref{fig:evaluation_bias}). Four of five evaluators rated their own model's responses more favorably than did out-of-family evaluators, with differences from $+7.91$ to $+16.29$ percentage points; Qwen3-4B was the exception ($-4.42$ percentage points). Five of six same-family comparisons were also positive ($+4.00$ to $+11.54$ percentage points), whereas Gemini 2.5 Flash-Lite rated Gemini 2.5 Flash responses less favorably ($-6.17$ percentage points). All 11 bootstrap 95\% confidence intervals excluded zero.

\clearpage
\subsection*{Physician and \ac{LLM} Abstention (RQ5)}

Physician abstention increased monotonically with physician-rated difficulty, with the median per-physician rate rising from 2.68\% in the lowest tier to 13.74\% in the highest (Table~\ref{tab:abstention_difficulty}). Most \ac{LLM} evaluators showed little or no increase across difficulty tiers. GPT-5 Nano was an exception, with rates of 0.08\%, 0.08\%, 0.21\%, and 1.38\% across the four tiers. Gemini 3 Flash and GPT-5.4 Mini did not abstain in any tier. Gemma 3 4B and Qwen3-4B had the highest global rates (6.41\% and 4.23\%), while each of the other seven evaluators had a global rate of at most 0.51\%.

\begin{table}[!ht]
\caption{Physician and large language model (LLM) evaluator abstention by physician-rated difficulty.}
\label{tab:abstention_difficulty}
\begin{adjustwidth}{-0.4in}{-0.4in}
\centering
\scriptsize
\setlength{\tabcolsep}{3pt}
\begin{tabular}{@{}lcccc@{}}
\toprule
Evaluator group & $\leq 0.5$ (Easy) & $0.5$--$1.0$ & $1.0$--$1.5$ & $>1.5$ (Hard) \\
\midrule
Physicians & 2.68 [0.81--4.12] & 4.73 [1.60--12.73] & 5.97 [2.93--33.04] & 13.74 [7.48--41.33] \\
\midrule
Gemini 3 Flash        & 0.00 (0)   & 0.00 (0)   & 0.00 (0)   & 0.00 (0) \\
GPT-5.4 Mini          & 0.00 (0)   & 0.00 (0)   & 0.00 (0)   & 0.00 (0) \\
Gemini 2.5 Flash-Lite & 0.08 (7)   & 0.00 (0)   & 0.03 (1)   & 0.00 (0) \\
Gemini 2.5 Flash      & 0.00 (0)   & 0.16 (10)  & 0.00 (0)   & 0.00 (0) \\
GPT-5.4 Nano          & 0.51 (46)  & 0.54 (34)  & 0.45 (13)  & 0.50 (4) \\
Gemma 3 27B           & 0.11 (10)  & 0.11 (7)   & 0.10 (3)   & 0.12 (1) \\
GPT-5 Nano            & 0.08 (7)   & 0.08 (5)   & 0.21 (6)   & 1.38 (11) \\
Qwen3-4B              & 4.86 (438) & 3.71 (235) & 3.57 (102) & 3.62 (29) \\
Gemma 3 4B            & 5.26 (474) & 7.65 (484) & 6.99 (200) & 7.50 (60) \\
\bottomrule
\end{tabular}
\end{adjustwidth}
\vspace{2pt}
\raggedright\scriptsize Values are percentages. Physician entries are median [interquartile range] across physician-specific abstention rates; the four tiers contained 23,630, 19,145, 8,070, and 2,120 physician ratings, respectively. \acs{LLM} entries are abstention rate (number of abstentions); each evaluator provided 9,010, 6,330, 2,860, and 800 ratings across the four tiers. Tiers were defined using mean physician-rated question difficulty.
\end{table}

\section*{Discussion}\label{sec:discussion}

\subsection*{Principal Findings}

MedQADE provides a physician-annotated benchmark for evaluating German open-response clinical answers. Across the five research questions, strong \ac{LLM} evaluators approached individual-physician agreement in this setting, while evaluator bias and low observed abstention raised additional evaluation concerns. The benchmark's heterogeneous student responses and physician-derived difficulty strata revealed both performance differences and evaluator behaviors that aggregate agreement alone would miss. Together, these findings motivate prospective assessment of physician-supervised annotation workflows, including evaluator provenance and the effectiveness of selective deferral.

\subsection*{Physician Reliability (RQ1)}

The physician panel provided a practical reference standard for an inherently interpretive annotation task. The correctness-agreement estimates (Table~\ref{tab:human_irr}), together with the high consensus coverage reported in the Results, show that multi-rater aggregation resolved most labels while retaining information about observed disagreement. Compared with relying on a single adjudicator, the combination of dense core ratings and targeted extension ratings reduced dependence on any one physician and offered a pragmatic strategy for constructing physician-anchored labels at scale.

The metric pattern also illustrates why agreement should not be summarized by a single statistic. For student models whose responses were predominantly judged incorrect (Gemma 3 4B and Qwen3-4B), high observed agreement coexisted with lower chance-corrected agreement because of label-prevalence imbalance. Reporting \ac{PA}, Cohen's $\kappa$, \ac{PABAK}, and Krippendorff's $\alpha$ therefore provided complementary views of panel consistency rather than competing conclusions.

The lower agreement on question difficulty means that the resulting tiers are best understood as aggregate physician perceptions rather than intrinsic item properties. They nevertheless provide a transparent exploratory stratification of how the panel collectively experienced the questions, without requiring the tiers to represent an immutable measure of clinical difficulty.

\subsection*{Student-Model Performance (RQ2)}

The broad performance spread across student models (Table~\ref{tab:student_lexical}) shows that MedQADE contains the response-quality variation needed to distinguish model capabilities and to test automated evaluators. Question-answer sets containing at least one exact lexical match were enriched for correct responses, but the models remained separable among question-answer sets for which no student response exactly matched the reference answer. Here, lexical match served as a scalable sampling and enrichment heuristic rather than as a correctness criterion. This finding complements prior evidence that automated metrics can capture human judgments more effectively than lexical metrics in medical open-response evaluation \cite{MORQA2025}.

The generally lower accuracy point estimates in higher physician-rated tiers (Figure~\ref{fig:student_difficulty}) are consistent with the aggregate ratings providing a behaviorally meaningful exploratory stratification. Lexical-match status and physician-rated difficulty provide complementary views of the benchmark: the former is an inexpensive pre-annotation signal, whereas the latter summarizes the panel's post-annotation assessment of question difficulty.

By including question-answer sets in which none of the generated answers exactly matched the reference answer, the benchmark focuses evaluation on responses for which expert assessment is most informative; its absolute accuracy estimates characterize this deliberately challenging evaluation set. The open-response format also removes recognition cues supplied by answer options, a distinction that can affect model performance \cite{Singh2025,Cocchieri2026}. Isolating that format effect within MedQADE would require a matched comparison with the same questions presented in multiple-choice form and is separate from the present model comparisons.

\subsection*{\ac{LLM} Evaluator Alignment and Ensembles (RQ3)}

The strongest \ac{LLM} evaluators (Gemini 3 Flash, GPT-5.4 Mini) were close to the physician reference (Table~\ref{tab:llm_alignment}), consistent with prior reports of strong agreement between automated judges and clinical experts \cite{Croxford2025,Chen2025,Zhang2025}. This makes them candidates for physician-supervised first-pass annotators or quality-control tools, where automation could accelerate evaluation while physicians remain the reference for adjudication and periodic validation.


The selected ensemble's small point-estimate gain over Gemini 3 Flash had a wide confidence interval spanning zero and was estimated after in-sample selection. A single strong evaluator may therefore offer a simpler and more resource-efficient workflow than coordinating several weaker evaluators in this setting. Held-out comparisons are needed to determine whether aggregation improves alignment.

\subsection*{Implications of Evaluator Bias (RQ4)}

The paired analysis identified these biases (Figure~\ref{fig:evaluation_bias}). These findings are consistent with previously reported self-preference in \ac{LLM}-based evaluation \cite{Bowman2024,Ackerman2024}. Because each target evaluator was compared with out-of-family evaluators judging the same responses, the observed differences cannot be attributed to variation in question composition or student-response quality between comparison sets. The bias was not uniformly positive: Qwen3-4B rated its own responses less favorably than out-of-family evaluators, and Gemini 2.5 Flash-Lite rated Gemini 2.5 Flash responses less favorably.

This dependence on evaluator identity has direct implications for model comparison. When student-model scores are close, differences of the observed magnitude could influence rankings or apparent performance differences. Automated benchmarks should therefore disclose evaluator identity, separate student and evaluator model families where feasible, and validate ratings against an independent physician-labeled subset. Multi-family evaluator panels may further reduce reliance on the scoring tendencies of any single model family.

\subsection*{Physician and \ac{LLM} Abstention (RQ5)}

The joint analysis reveals differences in abstention behavior between physicians and \acp{LLM} (Table~\ref{tab:abstention_difficulty}). Physician abstention increased with difficulty, consistent with the rubric's instruction to abstain when medical expertise was insufficient for a confident judgment. The two evaluators closest to the physician reference never abstained. GPT-5 Nano's abstention rates were nondecreasing across tiers, but this pattern was not shared consistently across the evaluator panel. Thus, the observed association between abstention and physician-rated difficulty was model-specific rather than uniformly absent.

The physician entries summarize distributions of per-rater rates, whereas each \ac{LLM} entry is a pooled model-specific rate. The comparison consequently characterizes observable label use rather than providing a matched test of human and model uncertainty calibration. Nevertheless, using the same question tiers supports the central operational observation that physicians and \acp{LLM} used the available abstention label differently.

Low abstention raises an operational concern for automated medical-evaluation pipelines: an incorrect definitive label does not itself provide an abstention-based routing signal for physician review. This observation is relevant to concerns about \ac{LLM} failure to abstain under clinical uncertainty \cite{Cocchieri2026Abstain}, but abstention frequency alone does not establish whether deferral would effectively identify likely errors. The conclusion concerns observed behavior rather than an internal psychological state: the study establishes how often the abstention label was used, not whether models represented uncertainty internally or translated it appropriately into deferral.

Automated evaluators should therefore be assessed on agreement and coverage jointly. Risk--coverage curves, calibrated deferral thresholds, error enrichment among abstained cases, and the downstream benefit of physician review would establish whether abstention functions as an effective safety mechanism. The objective is not indiscriminately higher abstention, but selective deferral of likely errors while preserving useful coverage.

\subsection*{Limitations}

First, MedQADE was derived from a single German medical-education source and was restricted to examination-style questions with one cloze deletion. The findings may not generalize to longer free-text answers, clinical dialogue, other languages or specialties, or documentation and decision-making in clinical practice. The study evaluated neither patient data nor patient outcomes; consequently, agreement with the physician panel should not be interpreted as evidence of clinical safety or deployment readiness.

Second, the 20:80 lexical-match sampling deliberately enriched the benchmark for responses anticipated to be difficult and therefore does not estimate performance on the complete 26,598-item source pool or on a natural distribution of clinical cases. The difficulty tiers were based on subjective physician ratings with low agreement and should be interpreted as exploratory aggregate perceptions rather than intrinsic item properties. Although the physician panel provided expert review, it was not sampled to represent every specialty or practice setting, and the split-annotation design assigned two primary raters per extension item, with the tenth physician providing an additional rating when consensus was absent. The resulting consensus may not capture every clinically acceptable interpretation.

Third, evaluator alignment and the observed biases are specific to the tested model versions, providers, prompts, and access periods and may change as proprietary systems are updated. The ensemble search selected and evaluated candidates on the same 200-item core subset, making its results exploratory and potentially optimistic, and evaluator alignment was not validated on an external benchmark. Confidence-interval overlap was descriptive rather than a test of equivalence.

Finally, abstention was elicited by an expertise-based rubric rather than a calibrated error-risk threshold. Output-label frequencies do not reveal whether models represented uncertainty internally, whether their judgments were calibrated, or whether abstention selectively identified likely errors. Held-out and external validation, broader physician panels, and prospective assessment of risk--coverage behavior are needed before applying these evaluators beyond the present benchmark setting.

\subsection*{Conclusions}

MedQADE addresses the scarcity of expert-curated German medical \ac{QA} resources by providing a physician-annotated benchmark built from native German single-cloze questions and model-generated open responses in German. Rather than relying on translated or multiple-choice material, it enables direct evaluation of medical answer generation and automated grading in German. Its multi-rater design yielded a practical consensus reference, while the heterogeneous response set differentiated student models and automated evaluators. The strongest \ac{LLM} evaluators achieved agreement profiles descriptively close to individual-physician agreement, supporting their prospective evaluation in physician-supervised annotation workflows.

Dependable automated evaluation, however, requires more than high aggregate agreement. Paired comparisons identified evaluator bias, with more favorable ratings in most, but not all, comparisons. Abstention was rare in most evaluators; GPT-5 Nano showed a nondecreasing pattern with physician-rated difficulty, whereas the two evaluators closest to the physician reference never abstained. These findings motivate attention to evaluator provenance and further testing of selective deferral rather than establishing its effectiveness.

Together, the results support physician-anchored, \ac{LLM}-assisted evaluation in the tested setting of model-generated German answers to single-cloze clinical questions. Automated evaluators should be selected independently of the models being assessed where feasible, periodically validated against physician-labeled data, and assessed jointly for agreement and coverage. MedQADE provides a physician-annotated framework for comparing medical \acp{LLM} and automated evaluators directly on native German questions without depending on translated benchmarks.

\section*{Acknowledgments}
\noindent No additional contributors beyond the authors are acknowledged.
\noindent Use of Generative AI. GPT-5.6 Sol assisted with Figure 1 and manuscript writing, GPT-5.6 Terra with literature searches \cite{OpenAI2026GPT56}, and DeepSeek V4 Flash with code implementation and review \cite{DeepSeekAI2026V4}. The authors reviewed all AI-assisted outputs and take responsibility for the final content.

\section*{Funding}
\noindent This work was supported by BMFTR grant agreement no. 16SV8958 (SEPE). The funder had no role in study design, data collection, analysis, interpretation, manuscript writing, or the publication decision.

\section*{Data Availability}
\noindent The source questions were derived from the Ankizin v5 resources \cite{Ankizin2024,AnkizinLeitfaden2024}. The complete question text, physician annotations, generated answers, reference labels, evaluator outputs, prompts and configurations, and analysis code have been deposited in an embargoed Zenodo archive (DOI: 10.5281/zenodo.21722212). Full redistribution rights for the question text have been confirmed. The archived materials will be released publicly upon publication under the CC BY-NC 4.0 license. During peer review, editors and reviewers can obtain credentialed access to the embargoed archive upon request to the corresponding author.

\section*{Authors' Contributions}
\noindent Conceptualization: William Philipp. Methodology: William Philipp. Software: William Philipp, Finn Fassbender. Formal analysis: William Philipp. Data curation: Finn Fassbender, Martje G. Pauly, Rebecca Herzog, Markus A. Hobert, Theresa Paulus, Alexander Baumann, Chi Wang Ip, Lukas L. Goede, Johanna Engelhard, Sebastian Loens, Thorsten Langer. Investigation: Martje G. Pauly, Rebecca Herzog, Markus A. Hobert, Theresa Paulus, Alexander Baumann, Chi Wang Ip, Lukas L. Goede, Johanna Engelhard, Sebastian Loens, Thorsten Langer. Resources: Daniel Fister. Writing---original draft: William Philipp. Writing---review \& editing: Sebastian Fudickar, Finn Fassbender, Ronald B\"ock. Supervision: Sebastian Fudickar. Funding acquisition: Sebastian Fudickar. All authors approved the final manuscript.

\section*{Conflicts of Interest}
\noindent None declared.

\section*{Abbreviations}
\noindent AI: artificial intelligence; API: application programming interface; BMFTR: German Federal Ministry of Research, Technology and Space; CI: confidence interval; EU: European Union; GBS: Guillain-Barr\'{e} syndrome; JSON: JavaScript Object Notation; LLM: large language model; MDR: Medical Device Regulation; MMLU: Massive Multitask Language Understanding; NLP: natural language processing; PA: percent agreement; PABAK: prevalence-adjusted bias-adjusted kappa; QA: question answering; RQ: research question.

\section*{Multimedia Appendices}
\setlength{\parskip}{6pt}

\phantomsection
\paragraph*{Multimedia Appendix 1. Model Configuration and Execution}
\label{S1_Configuration}
\noindent Student-model roster and deployment. The five student models served as proxies for medical students: Gemini 2.5 Flash, GPT-5 Nano, Gemma 3 27B, Gemma 3 4B, and Qwen3-4B.

\noindent The corresponding implementation identifiers, in the roster order above, were \texttt{gemini-2.5-flash}, \texttt{gpt-5-nano}, \texttt{gemma3:27b}, \texttt{gemma3:4b}, and \texttt{qwen3:4b}, respectively.

\noindent The proprietary student models were accessed through their official APIs, whereas Gemma 3 27B, Gemma 3 4B, and Qwen3-4B were run locally using Ollama.

\noindent Approximate recorded student-run completion times (YYYY-MM-DD HH:MM) were Gemini 2.5 Flash: 2025-12-07 17:58; Gemma 3 4B: 2025-12-11 14:06; Qwen3-4B: 2025-12-15 15:35; GPT-5 Nano: 2025-12-17 10:07; and Gemma 3 27B: 2025-12-17 10:48.

\noindent Student execution settings. Student generation used temperature 0.2 for Gemini 2.5 Flash, Gemma 3 27B, Gemma 3 4B, and Qwen3-4B; GPT-5 Nano used the provider's default temperature because user-defined values were not supported. GPT-5 Nano used \texttt{low} reasoning effort, Gemini 2.5 Flash used a thinking budget of 0, and thinking was disabled for the Ollama-hosted models using \texttt{think=false}.

\noindent Evaluator roster and identifiers. The same five models were also used as automated evaluators, and the four added evaluators were Gemini 3 Flash, Gemini 2.5 Flash-Lite, GPT-5.4 Nano, and GPT-5.4 Mini.

\noindent The five shared models use the identifiers given above. Gemini 3 Flash used \texttt{gemini-3-flash-preview}, requested through the provider alias \texttt{gemini-flash-latest}; Gemini 2.5 Flash-Lite used \texttt{gemini-2.5-flash-lite}, requested through \texttt{gemini-flash-lite-latest}. GPT-5.4 Nano and GPT-5.4 Mini used \texttt{gpt-5.4-nano} and \texttt{gpt-5.4-mini}, respectively.

\noindent Evaluator execution settings. Evaluator temperature was 0 where supported, with the provider's default used for GPT-5 Nano. GPT-5 Nano, GPT-5.4 Nano, and GPT-5.4 Mini used reasoning effort \texttt{none}; Gemini 2.5 Flash used a thinking budget of 256 tokens; Gemini 3 Flash used the \texttt{minimal} thinking level, the lowest supported setting; Gemini 2.5 Flash-Lite had no explicit thinking configuration; and thinking was disabled for the Ollama-hosted models using \texttt{think=false}.

\noindent Approximate recorded evaluator-run completion times (YYYY-MM-DD HH:MM) were GPT-5 Nano: 2026-02-16 18:53; Gemma 3 4B: 2026-02-17 14:01; Qwen3-4B: 2026-02-17 23:16; Gemini 3 Flash: 2026-03-13 21:48; Gemini 2.5 Flash: 2026-03-16 12:07; Gemini 2.5 Flash-Lite: 2026-03-16 13:15; Gemma 3 27B: 2026-03-19 09:55; GPT-5.4 Nano: 2026-03-19 11:25; and GPT-5.4 Mini: 2026-03-19 12:36.

\noindent Evaluator output format. Evaluator responses were constrained using a JSON Schema containing the predefined correctness and difficulty labels, which facilitated automated parsing and prevented outputs outside the permitted label set.

\phantomsection
\paragraph*{Multimedia Appendix 2. Student-Model Prompt}
\label{S2_Appendix}
\noindent System prompt (German; used for generation). The following listing is the verbatim German system prompt used for all LLM student models to generate cloze-question answers:

\begin{lstlisting}
Du bist ein erfahrener Medizinstudent in einer anspruchsvollen, zeitkritischen Prüfung.

Deine Aufgabe ist es, eine medizinische Lückentext-Frage (Cloze-Frage) mit genau einer Lücke zu beantworten. Fülle diese einzelne Lücke fachlich korrekt aus.

Regeln für die Ausgabe:
1. Liefere ausschliesslich das fehlende Wort oder den fehlenden kurzen Fachbegriff als deine Antwort.
2. Verwende ausschließlich die korrekte deutsche medizinische Fachterminologie.
3. Gib genau eine präzise und spezifische Antwort.
4. Keine zusätzlichen Erklärungen, Sätze, Anführungszeichen, Nummern, Satzzeichen oder Formatierungen.

Beispiel:
Eingabe: Das Hormon, das den Blutzuckerspiegel senkt, ist ___.
Erwartete Ausgabe: Insulin
\end{lstlisting}

\noindent\textit{English translation (for readability; not supplied to models):}

\begin{lstlisting}
You are an experienced medical student in a demanding, time-critical exam.

Your task is to answer a medical cloze question with exactly one gap. Fill this single gap correctly.

Output rules:
1. Deliver only the missing word or short technical term as your answer.
2. Use only correct German medical terminology.
3. Give exactly one precise and specific answer.
4. No additional explanations, sentences, quotation marks, numbers, punctuation, or formatting.

Example:
Input: The hormone that lowers blood sugar is ___.
Expected output: Insulin
\end{lstlisting}

\phantomsection
\paragraph*{Multimedia Appendix 3. Human Annotation Guidelines and Evaluator Prompt}
\label{S3_Appendix}
\noindent System prompt (German; verbatim). The following German listing is the verbatim system prompt used for all LLM rater models to evaluate the correctness of student-generated answers and the difficulty of the corresponding questions. Apart from the role-prompting element, this prompt is identical to the annotation guidelines provided to the human expert panel. The difficulty rating was collected as part of the LLM annotation process but was not analyzed in this study.

\noindent User input. Each evaluation request included the clinical question, the original reference answer as context (not a grading key), and five student answers. The system instructed independent assessment of each answer.

\noindent Evaluator structured-output contract. The parser recorded the fields shown below. The angle-bracket notation denotes the permitted value set, not a literal output.

\begin{lstlisting}
{
  "resp_1": "<Richtig|Falsch|Keine Angabe>",
  "resp_2": "<Richtig|Falsch|Keine Angabe>",
  "resp_3": "<Richtig|Falsch|Keine Angabe>",
  "resp_4": "<Richtig|Falsch|Keine Angabe>",
  "resp_5": "<Richtig|Falsch|Keine Angabe>",
  "difficulty": "<Einfach|Mittel|Schwer>"
}
\end{lstlisting}

\begin{lstlisting}
# Rolle

Sie sind eine medizinisch fachkundige Person mit Erfahrung in der Bewertung medizinischer Prüfungsfragen.
Ihre Aufgabe ist es, mehrere generierte Antworten zu derselben medizinischen Prüfungsfrage fachlich korrekt, sachlich und konsistent anhand der untenstehenden Kriterien zu bewerten.

Jede Antwort ist einzeln anhand der Kriterien zu beurteilen.

# Anleitung zur Bewertung von medizinischen Prüfungsantworten

## Ziel

Zu jeder medizinischen Prüfungsfrage werden mehrere generierte Antworten vorgelegt.

Bewerten Sie jede einzelne Antwort anhand der folgenden Kategorien.

Eine Beispielantwort wird zur Orientierung bereitgestellt, dient jedoch nicht der Beurteilung der generierten Antworten selbst.

Auch wenn mehrere Antworten gleichzeitig vorliegen, ist jede Antwort für sich anhand der Kriterien zu bewerten.

## Bewertungskategorien

### 1. Richtig

Die Antwort ist inhaltlich korrekt, medizinisch präzise und beantwortet die Frage vollständig.

- Alle wesentlichen Punkte sind enthalten.
- Es sind keine medizinisch falschen oder irreführenden Aussagen enthalten.
- Kleinere Unterschiede in Formulierung oder Detailtiefe sind akzeptabel, solange die Aussage vollständig korrekt ist.
- Fachlich korrekte Synonyme oder spezifischere Bezeichnungen gelten als richtig.

### 2. Falsch

Die Antwort ist ganz oder teilweise inhaltlich falsch, unvollständig oder missverständlich, sodass sie keine korrekte Lösung darstellt.

Bewerten Sie auch folgende Fälle als falsch:

- Die Antwort nennt nur Teilaspekte, ohne die vollständige Lösung zu liefern.
- Die Antwort ist sehr unpräzise oder spekulativ.
- Die Antwort enthält medizinische Fehler.
- Die Antwort enthält potenziell gefährliche Aussagen.
- Die Antwort klingt plausibel, ist aber fachlich nicht korrekt.
- Eine notwendige Spezifikation fehlt.

### 3. Keine Angabe

Wählen Sie diese Option nur, wenn Sie nicht über die notwendige medizinische Expertise verfügen, um die Antwort fachlich sicher zu bewerten.

- Es geht nicht darum, ob die Antwort gut oder schlecht ist, sondern ausschließlich darum, dass die fachliche Beurteilung nicht sicher getroffen werden kann.

## Beispiele zur Verdeutlichung

### Beispiel 1

Frage: Welches Vitamin ist wichtig für die Blutgerinnung?
Beispielantwort: Vitamin K

Generierte Antworten:

- Vitamin K -> Richtig
- Fettlösliche Vitamine -> Falsch (zu allgemein, nur ein fettlösliches Vitamin ist korrekt)
- Phyllochinon -> Richtig (chemischer Name von Vitamin K, korrekt)

### Beispiel 2

Frage: Welches Organ produziert Insulin?
Beispielantwort: Bauchspeicheldrüse

Generierte Antworten:

- Bauchspeicheldrüse -> Richtig
- Pankreas Langerhans-Inseln -> Richtig (sehr spezifisch, korrekt)
- Leber -> Falsch (falsches Organ)
- Drüse im Bauchraum -> Falsch (zu vage, beschreibt nur allgemein)

### Beispiel 3

Frage: Welche Blutgruppe gilt als Universalspender?
Beispielantwort: 0 negativ

Generierte Antworten:

- 0 negativ -> Richtig
- 0 -> Falsch (zu vage, muss negativ sein)
- 0 Rh-negativ -> Richtig (präzise, korrekt)

## Einschätzung der Schwierigkeit

Zusätzlich ist die Schwierigkeit der Frage nach eigenem Empfinden zu bewerten.

Optionen:

- Einfach
- Mittel
- Schwer

Diese Einschätzung ist subjektiv und dient ausschließlich der statistischen Auswertung der Fragen, nicht der Bewertung der einzelnen Antworten.
\end{lstlisting}

\noindent\textit{English translation (for readability; not supplied to models):}

\begin{lstlisting}
# Role

You are a medically knowledgeable person with experience in evaluating medical exam questions. Your task is to professionally, objectively, and consistently evaluate multiple generated answers to the same medical exam question according to the criteria below. Each answer is to be assessed individually against the criteria.

# Instructions for Evaluating Medical Exam Answers

## Objective

Multiple generated answers are presented for each medical exam question. Evaluate each individual answer according to the following categories. A sample answer is provided for orientation but is not to be used for judging the generated answers themselves. Even when multiple answers are presented simultaneously, each answer is to be evaluated independently against the criteria.

## Evaluation Categories

### 1. Correct

The answer is substantively correct, medically precise, and fully addresses the question.

- All essential points are included.
- No medically incorrect or misleading statements are present.
- Minor differences in phrasing or level of detail are acceptable as long as the statement is entirely correct.
- Medically correct synonyms or more specific terms are considered correct.

### 2. Incorrect

The answer is wholly or partially incorrect, incomplete, or ambiguous such that it does not constitute a correct solution. Also evaluate the following cases as incorrect:

- The answer addresses only partial aspects without providing the complete solution.
- The answer is very imprecise or speculative.
- The answer contains medical errors.
- The answer contains potentially dangerous statements.
- The answer sounds plausible but is not technically correct.
- A necessary specification is missing.

### 3. Abstain

Select this option only if you lack the necessary medical expertise to confidently evaluate the answer.

- This is not about whether the answer is good or bad, but solely about whether a confident professional assessment cannot be made.

## Illustrative Examples

### Example 1

Question: Which vitamin is important for blood clotting?
Sample answer: Vitamin K

Generated answers:

- Vitamin K -> Correct
- Fat-soluble vitamins -> Incorrect (too general, only one fat-soluble vitamin is correct)
- Phylloquinone -> Correct (chemical name of Vitamin K, correct)

### Example 2

Question: Which organ produces insulin?
Sample answer: Pancreas

Generated answers:

- Pancreas -> Correct
- Pancreatic islets of Langerhans -> Correct (very specific, correct)
- Liver -> Incorrect (wrong organ)
- Gland in the abdominal cavity -> Incorrect (too vague, only describes generally)

### Example 3

Question: Which blood type is considered the universal donor?
Sample answer: 0 negative

Generated answers:

- 0 negative -> Correct
- 0 -> Incorrect (too vague, must be negative)
- 0 Rh-negative -> Correct (precise, correct)

## Difficulty Assessment

Additionally, assess the difficulty of the question based on your own judgement.

Options:

- Easy
- Medium
- Hard

This assessment is subjective and is used exclusively for statistical analysis of the questions, not for evaluating the individual answers.
\end{lstlisting}

\end{document}